\documentclass[journal]{IEEEtran}
\usepackage{amsmath,graphicx}
\usepackage[caption=false,font=footnotesize]{subfig}
\usepackage{amsthm}
\usepackage{array}
\usepackage{float}
\newcolumntype{P}[1]{>{\centering\arraybackslash}p{#1}}
\newcolumntype{M}[1]{>{\centering\arraybackslash}m{#1}}
\newtheorem{theorem}{Theorem}
\usepackage{cite}
\usepackage{amsmath,amssymb,amsfonts}
\usepackage{bm}
\usepackage{graphicx}
\usepackage{textcomp}
\usepackage{algorithm}
\usepackage{algpseudocode}

\newtheorem{definition}{Definition}
\def\BibTeX{{\rm B\kern-.05em{\sc i\kern-.025em b}\kern-.08em
		T\kern-.1667em\lower.7ex\hbox{E}\kern-.125emX}}

\ifCLASSINFOpdf
\else
\fi

\makeatletter
\def\endthebibliography{%
	\def\@noitemerr{\@latex@warning{Empty `thebibliography' environment}}%
	\endlist
}
\makeatother
\begin{document}

\title{Robust Group Subspace Recovery: A New Approach for Multi-Modality Data Fusion}

\author{Sally~Ghanem,~\IEEEmembership{Student Member,~IEEE}
	Ashkan~Panahi,~\IEEEmembership{Member,~IEEE,}
       Hamid~Krim,~\IEEEmembership{Fellow,~IEEE,}
        and~Ryan~A.~Kerekes,~\IEEEmembership{Member,~IEEE}
\thanks{This manuscript has been authored by UT-Battelle, LLC, under contract DE-AC05-00OR22725 with the US Department of Energy (DOE). The US government retains and the publisher, by accepting the article for publication, acknowledges that the US government retains a nonexclusive, paid-up, irrevocable, worldwide license to publish or reproduce the published form of this manuscript, or allow others to do so, for US government purposes. DOE will provide public access to these results of federally sponsored research in accordance with the DOE Public Access Plan (http://energy.gov/downloads/doe-public-access-plan).

An earlier version of this paper was presented at the 26th European Signal Processing Conference (EUSIPCO) in 2018 and was published in its Proceedings (https://ieeexplore.ieee.org/abstract/document/8553445).

This paper have been submitted and accepted in IEEE Sensors Journal. \textcopyright 2020 IEEE.  Personal use of this material is permitted.  Permission from IEEE must be obtained for all other uses, in any current or future media, including reprinting/republishing this material for advertising or promotional purposes, creating new collective works, for resale or redistribution to servers or lists, or reuse of any copyrighted component of this work in other works 

Sally Ghanem, Ashkan Panahi and Hamid Krim are with the Department
of Electrical and Computer Engineering, North Carolina State University, Raleigh,
NC, 27606, USA

Ryan A. Kerekes is with Oak Ridge National Laboratory, Oak Ridge, TN 37830 USA}}

\maketitle

\begin{abstract}
Robust Subspace Recovery (RoSuRe) algorithm was recently introduced as a principled and numerically efficient algorithm that unfolds underlying Unions of Subspaces (UoS) structure, present in the data. The union of Subspaces (UoS) is capable of identifying more complex trends in data sets than simple linear models. We build on and extend RoSuRe to prospect the structure of different data modalities individually. We propose a novel multi-modal data fusion approach based on group sparsity which we refer to as Robust Group Subspace Recovery (RoGSuRe).  Relying on a bi-sparsity pursuit paradigm and non-smooth optimization techniques, the introduced framework learns a new joint representation of the time series from different data modalities, respecting an underlying UoS model. We subsequently integrate the obtained structures to form a unified subspace structure. The proposed approach exploits the structural dependencies between the different modalities data to cluster the associated target objects. The resulting fusion of the unlabeled sensors' data from experiments on audio and magnetic data has shown that our method is competitive with other state of the art subspace clustering methods. The resulting UoS structure is employed to classify newly observed data points, highlighting the abstraction capacity of the proposed method.
\end{abstract}


\IEEEpeerreviewmaketitle

\section{Introduction}
 \label{sec:intro}

\IEEEPARstart{U}{nion} of Subspaces (UoS) is a novel approach for identifying
complex trends in datasets relative to simple linear models, like Robust Principal Component Analysis \cite{b38}. High dimensional data is rich with common and related features lying in corresponding subspaces and also other nonconforming structures which may be errors or outlier sparse structures. Sparse modeling has been broadly employed in the machine learning literature to model noisy data  \cite{b12}. Subspace clustering can be utilized to group data points picked from a union of low-dimensional subspaces in an unsupervised fashion. Subspace clustering has become a highly researched topic in computer vision on account of widely available and large amounts of visual data, which has in turn, further motivated the development of such representations \cite{b40} \cite{b41}. Other applications of subspace clustering include image segmentation \cite{b46}, image compression \cite{b44} and motion segmentation \cite{b45}. Union of Subspaces has been also incorporated into analysis dictionary learning to boost the classification performance in supervised learning \cite{b56}. In this work, we build on the procedure studied in \cite{b16}, in which a bi-sparse model, known as Robust Subspace Recovery (RoSuRe) via Bi-sparsity Pursuit, is employed as a framework to recover the union of subspaces in the presence of sparse corruptions. The UoS structure is unveiled by pursuing sparse self-representation of the given data. 

We explore the flexibility of the UoS methodology for a new application: Identification and tracking of vehicles by way of multimodal data fusion. This application is similar to the emerging challenge of sensor networks and their role in advanced surveillance technology, as sensor networks can also use multimodal data fusion to obtain more information and insight from broad data sources \cite{b47}. Multimodal data have increasingly become more accessible with the proposed intelligent transportation systems to gather more comprehensive and complementary information, which in turn, require optimized exploitation strategies taking advantage of system redundancy. While data fusion is not new, it has recently garnered a significant increase in research interest for the need of further foundational principles, and for the varied applications including image fusion \cite{b50} \cite{b51}, target recognition \cite{b52}, speaker recognition \cite{b54} and handwriting analysis \cite{b53}. Recent developments in sensor technology have provided more potential in developing real-time transportation systems technologies.  Traffic flow optimization, dynamic traffic management solutions, vehicle counting, travel time estimation, and other traffic modeling studies frequently require classification and identification of streams of vehicles. Moreover, accurate estimation of traffic parameters needs to be performed in real time for decision makers \cite{b2}. Conventional vehicle identification methods such as license plate recognition and Radio Frequency Identification Tags (RFID) have long been widely used for that purpose \cite{b4}. Unfortunately, such image-based methods are not always appropriate, for example, in applications that require low-power, low-cost, and robustness to environmental changes. Additionally, attention to privacy issues has raised more concerns over image acquisition. In contrast, further alternatives such as magnetic sensors and microphones are inexpensive and obviate privacy concerns \cite{b6}. Vehicles are primarily of metallic structure that perturb the earth's magnetic field, and hence produce unique magnetic signatures that have served to discriminate between vehicles\cite{b7}\cite{b8}. Audio sensors have also been extensively employed in the area of vehicle classification for different applications, and have proven their effectiveness and robustness\cite{b9} \cite{b10}. 

The objective of this work is to develop a multi-modal classification approach which we validate on vehicle classification and identification. The sensor network can either include similar or dissimilar sensors. Similar sensor fusion, which is the case when multiple sensors explore the same features, has been studied in \cite{b59} and \cite{b60}. On the other hand, dissimilar sensor fusion \cite{b61} employ different sensors to prospect different features along various dimensions of the target. Our framework assumes feature fusion from heterogeneous data modalities. However, the proposed approach is principled and sufficiently general, as no specific assumptions were made, thus making it equally applicable for homogeneous or heterogeneous data modalities. Sensor fusion of heterogeneous data has long been of interest since it can explore different characteristics of a target and provide information reinforcement for an increased resolvability of object identity. This additional information from multiple modalities enrich target characterization and support decision making. A comprehensive survey for data fusion is provided in \cite{b57} and \cite{b58}. 

In this work, we consider an operationally relevant unsupervised learning scenario, where no training dataset is provided and we adopt a data driven approach to determine vehicle signatures utilizing key features extracted from each sensor modality. We subsequently combine the features from each sensor modality to generate a desirable universal feature and increase the classification rate of specific vehicle classes. We employ two approaches for data fusion. The first approach relies on the bi-sparsity framework to recover the underlying subspace structure in each sensor modality separately and obtain a finer level of classification by combining them through addition, which we presented before in \cite{b37}. The second approach relies on robust group sparsity, where the data from different modalities are jointly exploited to obtain a unified common subspace structure in one step. The idea of group sparsity was previously utilized in \cite{b62}. The Multi-task Low-rank Affinity Pursuit (MLAP) was proposed to boost region-based image segmentation by fusing different image features which have to have the same dimensionality and can not be easily modified to address the present issue. On the other hand, our approach can address data modalities that have different dimensionalities and potentially incompatible in nature. MLAP was also based on LRR \cite{b39} while our approach is based on RoSuRe \cite{b16}. For fairness, we will compare our approach to other popular multimodal subspace clustering baselines such as LTMSC\cite{b64}, kMLRR \cite{b65}, kMSSC \cite{b65}, MSSC \cite{b65} and MLRR \cite{b65}. LTMSC \cite{b64} exploits the complementary information of different views by jointly employing tensors to explore the high order correlations. The other methods were introduced in \cite{b65} as multimodal extension for Sparse Subspace Clustering (SSC) \cite{b63} and Low-rank Representation-based (LRR) \cite{b39} by enforcing a common representation across data modalities.

The paper is organized as follows. In Section 2, we provide the fundamental concepts of the two approaches RoSuRe and RoGSuRe. In Section 3, we introduce the different sensing modalities that will be utilized for experimentation along with the pre-processing, feature selection and extraction techniques. In Section 4, we present the experimental results of our approach, while Section 5 provides concluding remarks.

\section{Information subspace-based fusion}
\subsection{Problem Formulation}
	Consider a set of data realizations indexed by \textit{k = 1, 2, ..., n}. Furthermore, assume \textit{T} data modalities, indexed by \textit{t = 1, 2, 3,..., T}. Each data realization can be represented as a $m-$dimensional vector  $\mathbf{x_k}(t) \in \mathbb{R}^m $, where $\mathbf{X}(t)=[\mathbf{x}_1(t) \  \mathbf{x}_2(t) \ ... \ \mathbf{x}_n(t)]$. The goal is to partition a set of realizations into clusters whose respective measurements for each modality is well-represented by a low-dimensional subspace. Mathematically, this is tantamount to seeking a partitioning \{$X^1,X^2,...,X^P$\} of [$n$] observations/realizations. Also, $P$ is the number of clusters indexed by $I$, such that there exist linear subspaces $S^I(t) \subset \mathbb{R}^m$ with $\mathrm{dim}$($S^I(t)$) $ \ll m$,
	where $\mathbf{x}_k(t) \in S^I(t)$ for all $t \in [T]$.

\subsection{Robust subspace recovery via Bi-sparsity Pursuit: Fusion of Subspace Information}

 The Robust Subspace Recovery via Bi-Sparsity Pursuit (RoSuRe) introduced in \cite{b16} was originally proposed for uni-modal data. We present an overview for the RoSuRe structure, which our modified fusion approach exploits. We subsequently elaborate on the formulation of our problem and describe how we address the multi-modal data fusion. We assume that all data samples may be corrupted by additive sparse errors. Therefore, the UoS structure is often corrupted and each data sample deviates from its original subspace. Specifically, we consider a set of data samples $\mathbf{X}=[\mathbf{x}_1, \mathbf{x}_2, ..., \mathbf{x}_n] \in R^{m\times n}$, where $n$ corresponds to the number of realizations and $m$ specifies the number of variables or features in each realization. The columns of the matrix $\mathbf{X}$ may be partitioned such that each part $X^I$ is decomposed into a low dimensional subspace and a sparse outlier (e.g., non-conforming data).
\begin{equation}\label{eq:1} 
X^I=L^I+E^I,I=1,...,P,
\end{equation} 
where each $L^I$ serves as a single low dimensional subspace of the original data, and $L=[L^1|L^2|...|L^P]$ is the desired union of subspaces. Furthermore, the partition recovers the clustering of the original data samples corrupted by the error $E=[E^1|E^2|...|E^P]$. The objective of this approach is to simultaneously retrieve the subspaces and  the noiseless samples from the observed noisy data. The RoSuRe via Bi-Sparsity pursuit is based on the idea of self-representation. In other words, $\mathbf{l}_i$ can be represented by the other samples from the same subspace $S(\mathbf{l}_i)$.
\begin{equation}\label{eq:2} 
\mathbf{l_i}=\sum_{i\neq j,I_j\in S(I_i)} w_{ij} \mathbf{l_j}.
\end{equation} 
The above relation can be represented in a matrix form as follows,
\begin{equation}\label{eq:3} 
\mathbf{L}=\mathbf{L} \mathbf{W}.
\end{equation}
Under a suitable arrangement/permutation of the data realizations, the sparse coefficient matrix $\mathbf{W}$ is an $n \times n$ block-diagonal matrix with zero diagonals provided that each sample is represented by other samples only from the same subspace. More precisely, $W_{ij}=0$ whenever the indexes $i,j$ correspond to samples from different subspaces. As a result, the majority of the elements in $W$ is equal to zero. After further approximations and relaxations, the problem is formulated as follows,
\begin{equation}\label{eq:4} 
\begin{aligned}
\min \limits_{\mathbf{W},\mathbf{E},\mathbf{L}} \|\mathbf{W}\|_1+ \lambda \|\mathbf{E}\|_1, \\
\mathrm {s.t.} \ \mathbf{X}=\mathbf{L}+\mathbf{E}, \ \mathbf{L}=\mathbf{L}\mathbf{W}, \ W_{ii}=0.
\end{aligned}
\end{equation}
where $\| \|_1$ denotes the $l_1$ norm, i.e. the sum of absolute values of the argument. The minimum of Eqn.(4) is approximated through linearized Alternating Direction Method of Multipliers ADMM\cite{b17} and the sparsity of both $\mathbf{E}$ and $\mathbf{W}$ is traced until convergence. See \cite{b16} for more details.
\subsubsection{Finding clusters in data using $\mathbf{W}$} \mbox{}\\
The resulting $\mathbf{W}$ can be exploited to evaluate an affinity matrix. The affinity matrix is computed by,
\begin{equation}
\mathbf{A}=\mathbf{W}+\mathbf{W}^T.
\end{equation}
Subsequently, the spectral clustering method in \cite{b25} is utilized for data clustering. The method can be summarized as follows: a matrix $\mathbf{D}$ is defined to be a diagonal matrix  whose $i^{th}$ diagonal element is the degree of the $i^{th}$ node, $i.e.$ the sum of $i^{th}$ row in $\mathbf{A}$. The standard graph Laplacian matrix is next constructed as follows,
\begin{equation*}
\mathbf{G}=\mathbf{D}^{-1/2}\mathbf{A}\mathbf{D}^{-1/2}. 
\end{equation*}
Next, the eigenvectors $\mathbf{s}_1, \mathbf{s}_2, ..., \mathbf{s}_k$ of $\mathbf{G}$ corresponding to the largest $k$ eigenvalues are computed, where $k$ is the desired number of clusters. The matrix $\mathbf{S}=[\mathbf{s}_1 \mathbf{s}_2... \mathbf{s}_k]$ is then formed by stacking the eigenvectors in columns. Each row of $\mathbf{S}$ is a point in $\mathbb{R}^k$, k-means is then used to cluster the rows of $\mathbf{S}$. Finally, the original point $\mathbf{x}_i$ is assigned to cluster $j$ iff row $i$ of the matrix $\mathbf{S}$ was assigned to cluster $j$. 

\subsubsection{Multi-modal Subspace Recovery via RoSuRe}\mbox{}\\
As previously stated, RoSuRe does not support multimodal data since the algorithm needs to be applied on each data modality individually. To overcome this problem, we apply RoSuRe on each data modality and then we integrate the resulting sparse coefficient matrices $\mathbf{W}(t)$'s, for $t= 1,2,...,T$ modalities, through adding them as follows,
\begin{equation}
\mathbf{W}_{Total}=\sum_{t=1}^T \mathbf{W}(t).
\end{equation} 
By doing so, we are reinforcing the relation between data points that exist in all data modalities as reflected by the elements of $\mathbf{W}(t)$. While this may be justified as ensuring a cross-sensor consistency, we are also reducing the noise variance introduced by the outliers. A similar strategy was explored in community detection in Social Networks \cite{b31}, where an aggregation of multi-layer adjacency matrices was found to yield a better Signal to Noise ratio, and ultimately improved performance. In multi-layer networks, edges that exist in multiple layers, encode different but related relations among data points. The subspaces/clusters in our case, share model commonalities for given targets for which the relations among data observations are reflected by the sparse (non-zero) elements of $\mathbf{W}(t)$. We next introduce an alternative solution by a joint optimization framework over multiple data modalities at the same time, producing one common subspace structure instead of separately doing so with RoSuRe. 
\subsection{Robust Group Subspace Recovery-Driven Fusion}
 In this subsection, we introduce a novel approach based on RoSuRe which we naturally adapt to multimodal data. We define  $ \mathbf{\Omega} = \{ \mathbf{W} (t) \}_{t=1} ^T$, where $ \mathbf{W} (t)=[w_{kj}(t)]_{k,j}$. We define the group l-norm $\parallel \mathbf{\Omega} \parallel_{1,2}$ as:
\begin{equation}
\parallel \mathbf{\Omega} \parallel_{1,2}=\sum_{k,j}\sqrt{\sum_{t=1}^T w_{k,j}^2(t)}
\end{equation}
We introduce the following optimization as the joint Sparse Subspace Clustering (JSSC) framework with group sparsity,
\begin{equation}
\min\limits_{\mathbf{\Omega} \mid w_{kk}(t)=0} \parallel \Omega \parallel_{1,2}+ \rho \sum_{t=1}^T \parallel \mathbf{W} (t) \parallel_1  \mathrm{s.t.} \ \mathbf{X}(t)=\mathbf{X}(t)\mathbf{W}(t).
\end{equation}
This procedure is justified by the observation that when each entity is represented by the other ones in the same cluster, the inter-class terms $w_{kj}(t)$ are zero for every $t$, which implies that $\parallel \mathbf{\Omega}\parallel$ is group sparse along the $t$ dimension. Moreover, minimizing group l-norm promotes a group-sparse solution. Equation (8) is separable in $j$, where $j$ is the column index of $\mathbf{W}$ and can therefore be minimized for every $j$ and thus re-written as follows,

\begin{equation}
\begin{aligned}
\min \limits_{\mathbf{\Omega_j}} \sum_{k=1}^n \sqrt{\sum_{t=1}^{T} w_{k,j}^2(t)}+ \rho \sum_{k,t} | \mathbf{w}_{kj} (t) |\\ 
\mathrm{s.t.} \ \mathbf{x}_j (t)= \mathbf{X}(t)\mathbf{w}_j(t) \ \forall t \ w_{jj}(t)=0.
\end{aligned}
\end{equation}

where $\Omega_j=\{w_j(t)\}_{j=1}^{n}$ and $\Omega =\{\Omega_j\}_{j=1}^{n}$. In order to validate our approach, our goal would be to prove that $w_{kj}=0 \ \forall \ k \notin \ S_\alpha $ and $ \ j \in S_\alpha $, where $S_\alpha$ is the index of the subspace containing $x_j(t)$.
	
\subsubsection{Robust Group Subspace Recovery} \mbox{}\\
Similarly to RoSuRe, we propose a robust and non-convex version of the above formulation in Eqn. (8). We assume that the data matrices include non-conforming elements to assume the structure $\mathbf{X}(t)=\mathbf{L}(t)+\mathbf{E}(t)$, where the columns of $\mathbf{L}(t)$ reside on their corresponding subspaces and $\mathbf{E}(t)$ is a sparse error matrix. The optimization problem is then rewritten as,
\begin{equation}
\begin{aligned}
\min\limits_{\mathbf{\Omega} \mid w_{kk}(t)=0} \parallel \Omega \parallel_{1,2}+ \rho \sum_{t=1}^T \parallel \mathbf{W}(t) \parallel_1 + \lambda \sum_{t=1}^T \parallel \mathbf{E}(t) \parallel_1\\
\mathrm {s.t.} \ \mathbf{X}(t)=\mathbf{L}(t)+\mathbf{E}(t), \ \mathbf{L}(t)=\mathbf{L}(t)\mathbf{W}(t),
\end{aligned}  
\end{equation}
which can be approximately solved by a primal-dual method, with an appropriate introduction of an augmented Lagrange form. To proceed, first note that Eqn. (9) can be reduced to a two-variable problem by substituting  $\mathbf{L}(t)$ with $\mathbf{X}(t)-\mathbf{E}(t)$ and using $\mathbf{L}(t)=\mathbf{L}(t)\mathbf{W}(t)$. Assuming $T$ modalities, the Lagrangian objective functional now becomes:
\begin{equation}
\begin{aligned}
L(\mathbf{\Omega}, \mathbf{W}(t), \mathbf{E}(t), \mathbf{Y}(t), \mu)=\parallel \Omega \parallel_{1,2}+ \rho \sum_{t=1}^T \parallel \mathbf{W}(t) \parallel_1 + \\
\lambda \sum_{t=1}^T \parallel \mathbf{E}(t) \parallel_1+\sum_{t=1}^T<\mathbf{L}(t)\mathbf{W}(t)-\mathbf{L}(t),\mathbf{Y}(t)>\\
+\sum_{t=1}^T \frac{\mu}{2}\parallel\mathbf{L}(t)\mathbf{W}(t)-\mathbf{L}(t)\parallel_F^2,
\end{aligned} 
\end{equation}
where $\mathbf{Y}(t)$ is a matrix of Lagrange multipliers and $\mu$ is a constant. Let $\mathbf{\hat{W}}(t)=\mathbf{I}-\mathbf{W}(t)$, following the Chambolle-Pock algorithm for non-smooth primal dual algorithms, we update the following update rules for $\mathbf{W}(t)$ and $\mathbf{E}(t)$,
\begin{equation}
\begin{aligned}
\mathbf{W}_{k+1}(t)=\arg \min\limits_{\mathbf{W}(t)} \parallel \mathbf{\Omega} \parallel_{1,2}+\rho \parallel \mathbf{W}(t)\parallel_1 \\ + <\mathbf{L}_{k+1}(t)\mathbf{W}(t)-\mathbf{L}_{k+1}(t),\mathbf{Y}_k(t)>
+\\ \frac{\mu_k}{2}\parallel\mathbf{L}_{k+1}(t)\mathbf{W}(t)-\mathbf{L}_{k+1}(t)\parallel_F^2
\end{aligned}
\end{equation}

\begin{equation}
\begin{aligned}
\mathbf{E}_{k+1}(t)=\arg \min\limits_{\mathbf{E}(t)} \lambda \parallel \mathbf{E}(t) \parallel_1+ <\mathbf{L}(t)\mathbf{\hat{W}}_{k+1}(t),\mathbf{Y}_k(t)>\\
+ \frac{\mu_k}{2}\parallel\mathbf{L}_{k+1}(t)\mathbf{\hat{W}}_{k+1}(t)\parallel_F^2
\end{aligned}
\end{equation}
The linearized ADMM in \cite{b17} is used to approximate Eqn. (11) and (12) as follows,
\begin{equation*}
\mathbf{W}_k^+(t)= \mathrm{prox}_\frac{\rho}{\mu\eta_1}(\mathbf{W}_k(t)+\frac{\mathbf{L}^T_{k+1}(\mathbf{L}_{k+1} \hat{W}_k(t)-\frac{\mathbf{Y}_k(t)}{\mu_k})}{\eta_1})
\end{equation*}
\begin{equation*}
\mathbf{W}_{k+1}(t)=\gamma_\frac{\rho}{\mu\eta_1}(\mathbf{W}_k^+(t))
\end{equation*}
\begin{equation*}
\mathbf{E}_{k+1}(t)= \gamma_\frac{\lambda}{\mu\eta_2}(\mathbf{E}_k(t)+\frac{(\mathbf{L}_{k+1} \hat{W}_k(t)-\frac{\mathbf{Y}_k(t)}{\mu_k})\mathbf{\hat{W}}^T_{k+1}}{\eta_2}),
\end{equation*}
where $\mathrm{prox}_{\beta}(A_{i,j}(t))=A_{i,j}(t)*\max\{ (\sqrt{\sum_{t=1}^T A_{i,j}(t)^2} -\beta),0     \}/\sqrt{\sum_{t=1}^T A_{i,j}(t)^2}$ and $\gamma_\tau(B_{i,j})= sign(B_{i,j})*\max \{ (\mid B_{i,j}\mid-\tau),0 \}$. The Lagrange multipliers are updated as follows,

\begin{equation}
\mathbf{Y}_{k+1}(t)=\mathbf{Y}_k(t)+\mu_k(\mathbf{L}_{k+1}(t)\mathbf{W}_{k+1}(t)-\mathbf{L}_{k+1}(t))
\end{equation}
\begin{equation}
\mu_k=\epsilon \mu_k.
\end{equation}

A summary of the resulting algorithm is given in Algorithm 1.
\begin{algorithm}
	\caption{Robust Group Sparse Clustering}
	Initialize: $\mathbf{X}(t)$, $\rho$, $\lambda$, $\epsilon$, $\eta_1$ and $\eta_2$\\
	while not converged do\\
	$\mathbf{L_{k+1}}(t)=\mathbf{X}(t)-\mathbf{E_k}(t)$\\
	$\mathbf{W}_k^+(t)= prox_\frac{\rho}{\mu\eta_1}(\mathbf{W}_k(t)+\frac{\mathbf{L}^T_{k+1}(\mathbf{L}_{k+1} \mathbf{\hat{W}_k}(t)-\frac{\mathbf{Y}_k(t)}{\mu_k})}{\eta_1})$\\
	$\mathbf{W}_{k+1}(t)=\gamma_\frac{\rho}{\mu\eta_1}(\mathbf{W}_k^+(t))$\\
	$W_{ii}(t)=0$\\
	$\mathbf{\hat{W_{k+1}}}(t)=\mathbf{I}-\mathbf{W_{k+1}}(t)$\\
	$\mathbf{E}_{k+1}(t)= \gamma_\frac{\lambda}{\mu\eta_2}(\mathbf{E}_k(t)+\frac{(\mathbf{L}_{k+1} \mathbf{\hat{W}_k}(t)-\frac{\mathbf{Y}_k(t)}{\mu_k})\mathbf{\hat{W}}^T_{k+1}}{\eta_2})$\\
	$\mathbf{Y}_{k+1}(t)=\mathbf{Y}_k(t)+\mu_k(\mathbf{L}_{k+1}(t)\mathbf{W}_{k+1}(t)-\mathbf{L}_{k+1}(t))$\\
	$\mu_k=\epsilon \mu_k$\\
	end while
\end{algorithm}

\subsubsection{Theoretical Discussion}\
Let $\mathbf{X}(t)$ represent the dataset with unit length data from the $t^{th}$ modality for every $t$ $\in$ [T]. Moreover, $S(t)=S^1(t) \cup S^2(t) \cup ... S^P(t) $ is the union of subspaces of the underlying structure, where $S^k(t)$ denotes the $k^{th}$ subspace of the $t^{th}$ modality. We seek the partitioning ${X^1, X^2,...,X^P}$ of $[n]$ observations as elaborated in Section II. In the following, we state the theorem that will support our approach. We study the case for $\rho$=0 for the sake of clarity, and compare to the work in \cite{b32} for the deterministic model; in which the orientation of the subspaces as well as the distribution of the points in each subspace is deterministic. The same setting was considered by Elhamifar et. al in \cite{b34} and \cite{b35}. In particular, this theorem provides the required condition for the angle between subspaces to guarantee exact recovery. We show that our multi-modal approach provides looser and less restrictive bounds on the angle between the subspaces as compared to single-modal approach in \cite{b32}. More precisely, we prove that the RoGSuRe algorithm requires a smaller angle between different subspaces across all the modalities to guarantee their exact recovery, which explains the gain we achieve by leveraging multi-modal data fusion. Before proceeding, we will state some important definitions. 
\begin{definition}
	(Group Subspace Detection Property) The subspaces $\left \{S^I (t) \right \}_{I=1}^P$ and points $\mathbf{X}(t)$ obey the group subspace detection if and only if it holds that for all $i$, the optimal solution to Eqn. (9) has nonzero entries only when the corresponding columns of $\mathbf{X}(t)$ are in the same subspace as $\mathbf{x}_i(t)$.
\end{definition} 

\begin{definition}
	The inradius of a convex subset $P$ of a finite dimension Euclidean space, denoted by $r(P)$, is defined as the radius of the largest Euclidean ball inscribed in $P$.	
\end{definition}

\begin{definition}
We take $P_{-j}=\left \{  \{\xi (t) x_q(t)\}_{t} \ | \ \sum\limits_{t=1}^T\xi^2(t)\leq 1,\ q\neq j \right \}$, where $q$ belongs to the same subspace as $j$.
\end{definition}

\begin{theorem}
	Let $\mathbf{X}(t)$ represent the dataset with unit length data for every $t \in T$. Suppose that Eqn.(9) has a feasible solution $\Omega$ where $w_{ij}(t)=0$ for all $i,j$ not belonging to the same subspace. Let $\theta(t)$ be the smallest angle between vectors from distinct subspaces in the $t^\mathrm{th}$ modality.  For $\rho$=0, if $ \max_t \cos^2\theta (t) \textless \min_j r^2(P_{-j})$, then the subspace detection property holds.
\end{theorem}

The theorem basically guarantees that the subspace detection property holds as long as for any two subspaces across all data modalities, the minimum angle is less than the minimum inradius of $P_{-j}$ for all data modalities. It is easy to see that if the angle between a point on one subspace and an arbitrary direction on another (a dual direction) is small, these two subspaces will be close, hence, clustering becomes hard. Moreover, if the minimum inradius is small, which implies that the points are skewed towards specific direction and aren't well spread throughout the subspace, therefore, the clustering will also be difficult. In short, the theorem affirms that as long as different subspaces for all data modalities are not likewise oriented and the points on each subspace are sufficiently spread and diverse, RoGSuRe will successfully cluster the data. By comparing our results to Theorem 2.5 in \cite{b32}, it is easy to see that $P_{-j}$ for $t>1$ is much larger than the single modal approach in their paper and therefore the minimum inradius over $t$ is larger. As a result, the angle between two subpaces has a smaller upper bound for multimodal data as compared to the single modal case in \cite{b32}. The proof of Theorem 1 is presented in Appendix A.
  
\section{Multi-modal sensing in Vehicle Classification }
As previously stated, the goal is to integrate the union of subspaces structure underlying the data measurements from each sensor modality to support decision making. A roadside sensor system was configured to collect data from passing vehicles using various sensors, including a camera, microphone, laser range-finder, magnetometer, and low-frequency RF antenna. The same dataset was used in \cite{b36}. In this study, we are using the signatures captured using passive magnetic and acoustic sensors. The magnetic signatures are recorded using a single three-axis magnetic sensor, while the acoustic data is collected by a single microphone. The sensors were mounted on a rigid rack for ease of deployment and management. The data collection was conducted in a park environment, with limited public interference.

 The data is collected for seven different vehicles; two SUVs, one sedan and four trucks. The two SUVs are GMC Yukon and Hyundai Tucson, the sedan is Honda Accord. The four trucks are Chevrolet pickup truck, 14 ft rental moving truck and two Ford F-150s, one has a mounted top on the bed and the other one does not. The seven different vehicles were driven by the system yielding a total of 546 observations per sensor. Our goal is to analyze this dataset and distinguish seven classes where each class corresponds to one car. Furthermore, our goal is to be able to classify a newly observed dataset, using the structure learned through the current unlabeled data. For this purpose, the observations were divided into training and testing as discussed in Table 1. As shown in the table, we used 50 observations for each car in the learning phase, and the rest of the observations for validation. Sample outputs of the magnetometer and microphone are shown in Fig. 1(a) and 1(b) respectively.

\begin{table}[htbp]
	\caption{The dataset description}
	\begin{center}
		
		\begin{tabular}{|c|c|c|c|c|}
			\hline
			
			\ Vehicle & Training points & Testing points    \\
			\hline 
			\ Chevrolet Truck & 50 & 29 \\ 
			\hline
			\ Ford F-150 (Topper)  & 50 & 19  \\ 
			\hline 
			\ Ford F-150   & 50 & 31  \\ 
			\hline 
			\ GMC Yukon  & 50 & 24  \\ 
			\hline 
			\ Honda Accord  & 50 & 41  \\ 
			\hline 
			\ Hyundai Tuscon & 50 & 20  \\ 
			\hline 
			\ Uhaul Truck  & 50 & 32  \\ 
			\hline 
			
		\end{tabular}
		
		\label{tab1}
	\end{center}
\end{table}  
\begin{figure}[htb]
	\centering
	\subfloat[Sample output for magnetometer data.]{\includegraphics[width=4.4cm,height=3.5cm]{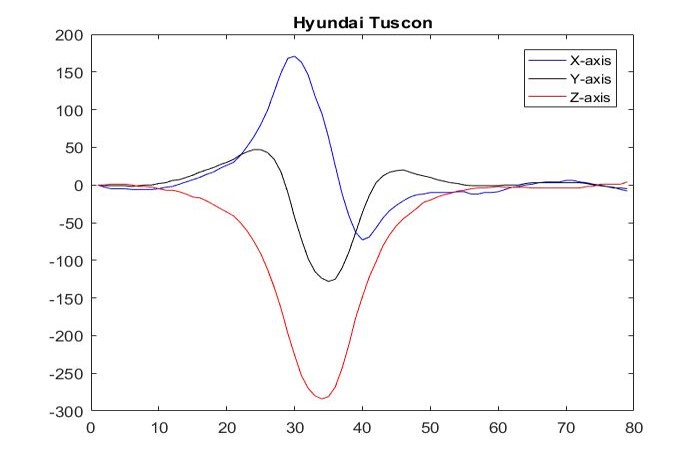}\label{fig:f11}}
	\hfill
	\subfloat[Sample output for audio data.]{\includegraphics[width=4.4cm,height=3.5cm]{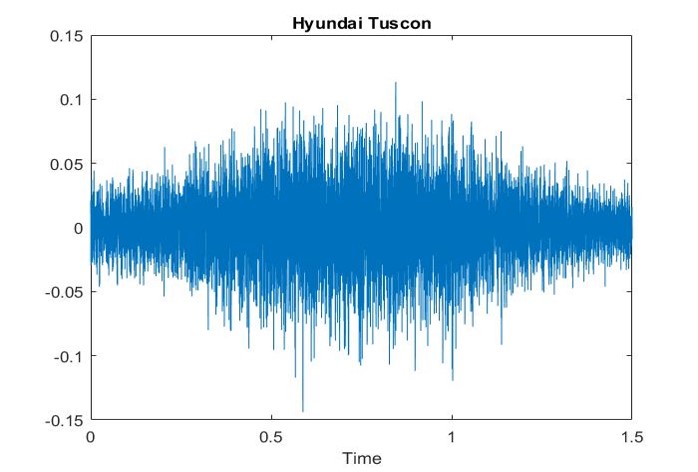}\label{fig:f12}}
		\caption{Sample input data from both sensors.}

\end{figure}

Acoustic sensors have been analyzed in various applications related to automatic transportation systems \cite{b18} \cite{b20}. Mel Frequency Cepstral Coefficients \cite{b21} are widely used in automatic speech recognition literature. They were introduced by Davis and Mermelstein in the 1980's, and have been used extensively to date. The Mel-Frequency Cepstrum (MFC) is a representation of the short-term power spectrum of a sound, based on a linear cosine transform of a log power spectrum on a nonlinear Mel scale of frequency. We extract and process MFCCs from our audio data. In our experiments, a low pass filter was applied to the audio signals to remove noise. Signals were downsampled from 192 kHz to 64 kHz. The audio signals were divided into windows of size 0.025 seconds with a step size of 0.01 seconds to allow some overlap between the frames, and thereby get a reliable spectral estimate. Finally, the MFFCs were extracted for each window and the highest 25 coefficients are selected to result in 2900 log filterbank energies for each observation. This setting achieved the highest classification performance in our experiments.

Magnetic sensors operate by detecting the variation in the magnetic inductance. Magnetic signatures can be characteristic of the vehicle of interest. Earth's magnetic field distortion can be used not only for the detection, but also for the classification and recognition of transport vehicles \cite{b23} \cite{b7} \cite{b8}. The exploited three-axis system is capable of producing up to 154 Hz and outputs 16-bit values with 67 Gauss resolution. In our experiment, a sample rate of 40 Hz has been used. For calibration, the magnetic signatures were extracted from the magnetic signals by subtracting the value of the local magnetic field, which is measured when no car passed by the sensor. The beginning and the ending of the signal are subsequently determined. Each observation is then normalized and re-sampled to get a normalized length of 100 samples per axis for a total of 300 samples per observation. The X, Y and Z signal amplitudes are re-scaled/normalized to fall in the [-1,1] interval.

\section{EXPERIMENTAL RESULTS } 

\subsection{RoSuRe-Based Fusion}

\subsubsection{Applying Uni-modal RoSuRe} \mbox{}\\
In the following, we use the RoSuRe technique to recover the subspace structure embedded in the data associated with each of the magnetic and audio sensors. The sparse solution of the problem in Eqn.(4), $\mathbf{W}(t)$,  provides important information about the relations among data points, which may be used to split data into individual clusters residing in a common subspace. Observations from each car can be seen as data points spanning one subspace.
We first proceed to extract the principal components of each of the sensor data\cite{b24}. The largest 100 principal values for magnetometer data and the largest 11 principal components for audio data are selected to serve as representatives of the data in the principal component space. Using the enhanced lower dimensional representation of the data, the sparse UoS coefficient matrix is obtained using RoSuRe by way of Eq. (4). The sparse coefficient matrix $\mathbf{W}(t)$ $\{ $$t=1$ for magnetometer, $t=2$ for audio $\}$ is thus computed using PCA-based representation in lieu of $\mathbf{X}(t)$. The affinity matrix is then calculated and the spectral clustering classification technique explained in II.B.1 is utilized to cluster the subspaces. The sparse coefficient matrices for magnetic and acoustic sensors are respectively illustrated in Fig. 2(a) and 2(b). The block-diagonal structure can be clearly seen from either of the matrices.  
\begin{figure}[htb]
	\centering
	\subfloat[The sparse coefficient matrix for magnetometer data]{\includegraphics[width=4cm,height=2.9cm]{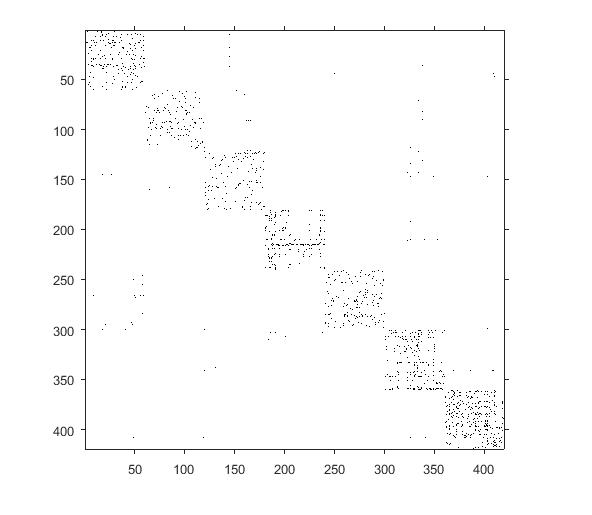}\label{fig:f21}}
	\hfill
	\subfloat[The sparse coefficient matrix for audio data]{\includegraphics[width=4.5cm,height=2.9cm]{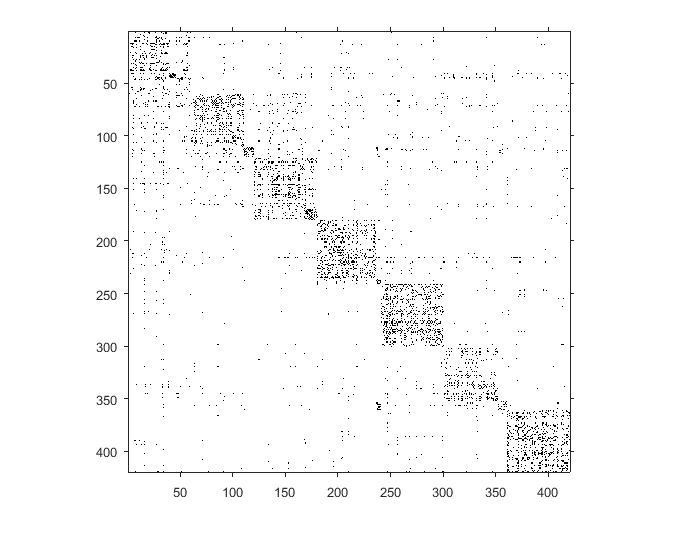}\label{fig:f22}}
	\centering
	\hfill
	\subfloat[The overall sparse coefficient matrix.]{\includegraphics[width=4cm,height=2.9cm]{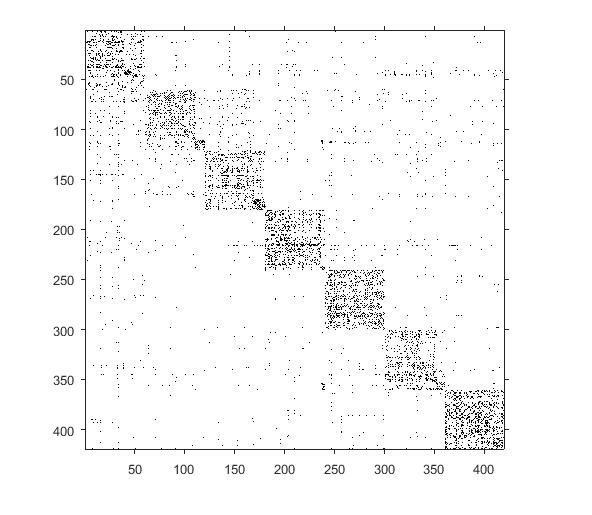}\label{fig:f23}}

	\caption{The sparse coefficient matrices for ROSURE.}
\end{figure}

\begin{table}[htbp]
	\caption{The clustering performance for different clustering methods}
	\begin{center}
		
		\begin{tabular}{|c|c|c|c|c|}
			\hline
			
			& RoSuRe & kmeans & GMM & HCA   \\
			
			\hline 
			
			Magnetomter data & 86.71\%  & 82.29\% & 77.14\% & 64.57\% \\ 
			\hline
			Audio data & 88.86\%  & 52.1\% & 62.57\%  & 40\% \\ 
			\hline 
			Fused & 98.29\%  & 82.29\% & 77.14\%  & 64.57\% \\ 
			\hline 
			
		\end{tabular}
		
		\label{tab2}
	\end{center}
\end{table}
\subsubsection{Bi-Modality RoSuRe-Driven Fusion (Multi-modal RoSuRe)} \mbox{}\\
Interpreting the subspace-based affinities based on $\mathbf{W}(t)$ as a layered set of networks, we proceed to carry out what amounts to modality fusion. The two sparse matrices $\mathbf{W}_{audio}$ and $\mathbf{W}_{magnetic}$ are added to produce one sparse matrix for both modalities, $\mathbf{W}_{Total}$, thereby improving performance. By doing so, we reinforce the contribution of similar representations that exist in both modalities as justified in II.B.2. The overall sparse matrix, $\mathbf{W}_{Total}$ is displayed in Fig. 2(c). Observations belonging to one car are clustered as one subspace where the contribution of each sensor is embedded in the entries of the $\mathbf{W}_{Total}$. For clustering by $\mathbf{W}_{Total}$, we applied the same spectral clustering approach that we previously demonstrated in II.B.1. As a result, the classification accuracy improved to 98.29\% as highlighted in Table 2. 

The performance of RoSuRe was compared against three widely used unsupervised clustering algorithms, namely, k-means, the Gaussian mixture model and hierarchical cluster analysis (HCA). k-means clustering, also referred to as the Lloyd-Forgy algorithm, is a computationally efficient method for cluster analysis in data mining \cite{b26}. k-means clustering aims to partition $n$ observations into $k$ clusters in which each observation belongs to the cluster with the nearest mean, serving as a representative of the cluster. A Gaussian mixture model is a probabilistic model which assumes all the data points generated from a mixture of a finite number of Gaussian distributions with unknown parameters. Mixture models can be considered as a generalization for k-means clustering to incorporate information about the covariance structure of the data as well as the centers of the latent Gaussians. Mixture models are in general less sensitive to the initialization of centroids. They have been used for feature extraction from speech data and object tracking \cite{b28} \cite{b29}. Hierarchical clustering is a technique which aims to build a hierarchy of clusters \cite{b30}. 

In our experiment,  we used a bottom-up approach where all observations start in their own cluster, pairs of clusters are subsequently merged together according to their closeness. The Euclidean distance, $d(\mathbf{x}_i,\mathbf{x}_j) = \Vert \mathbf{x}_i-\mathbf{x}_j \Vert_2$, was used as a proximity measure between each pair of data points. We used complete-linkage criterion to measure the distance between clusters where the distance $D(X,Y)$ between clusters $X$ and $Y$ is described as follows: $D(X,Y)= \max \limits_{\mathbf{x} \in X, \mathbf{y} \in Y} \ d(\mathbf{x},\mathbf{y}) $ The results are displayed in Table 2. As shown in the table, RoSuRe has the highest classification accuracy for both audio and magnetometer data. Moreover, after fusing the two data modalities, RoSuRe shows a significant enhancement in the classification performance. Additionally, we compared the RoSuRe fusion performance with the other unsupervised clustering methods through linking the two modalities features. More specifically, we concatenated both magnetometer and audio observations in one vector and we then clustered the new representation of the data. The results in Table 2 show that, by concatenating the data, we are not gaining extra information. Moreover, the classification accuracy after concatenation is the same as that of the magnetometer data because of the dominant higher dimensionality of magnetometer observations as compared to audio observations. The results were therefore biased towards the former modality. Whereas, by integrating the sparse coefficient matrix corresponding to each modality, we have obviously boosted the performance of RoSuRe from approximately 86\% to 98.29\%.

\subsection{Using Robust Group Subspace Recovery (RoGSuRe)}
\subsubsection{Fusing the data modalities with RoGSuRe} \mbox{}\\ 
In this subsection, we use the Robust Group Subspace Recovery technique to recover the subspace structure embedded in the data associated with each magnetic or audio observation. We follow the same data analysis explained in section III. We start by extracting the principal components of the data corresponding to each sensor\cite{b24} to serve as representatives of the data in the principal component space (in some sense denoised). The sparse coefficient matrix $\mathbf{W}(t)$ is computed by solving Eqn.(9) through Algorithm 1. Next, we threshold $\mathbf{W}(t)$ by its median value. The sparse coefficient matrices for magnetic and acoustic sensors are respectively illustrated in Fig. 3(a) and 3(b). The block-diagonal structure can be clearly seen from either of the matrices. 
\begin{figure}[htb]
	\centering
	\subfloat[The sparse coefficient matrix for magnetometer data $\mathbf{W}(1)$.]{\includegraphics[width=4cm,height=2.9cm]{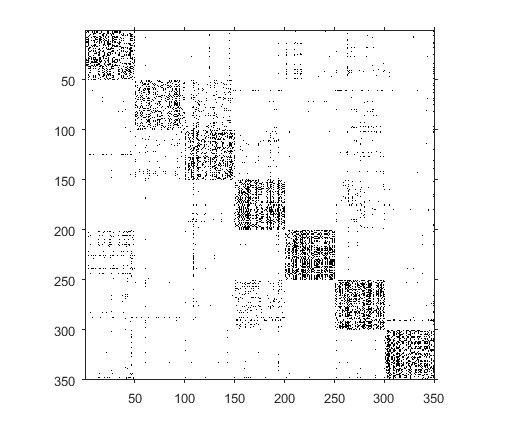}\label{fig:f30}}
	\hfill
	\subfloat[The sparse coefficient matrix for audio data $\mathbf{W}(2)$.]{\includegraphics[width=4.5cm,height=2.9cm]{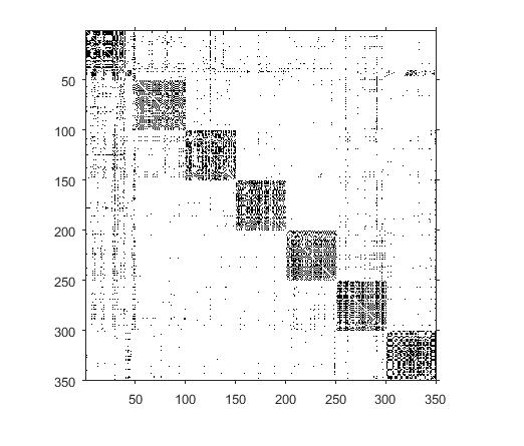}\label{fig:f31}}
	\centering
	\hfill
	\subfloat[The overall sparse coefficient matrix $\mathbf{W}_{total}$]{\includegraphics[width=4cm,height=2.9cm]{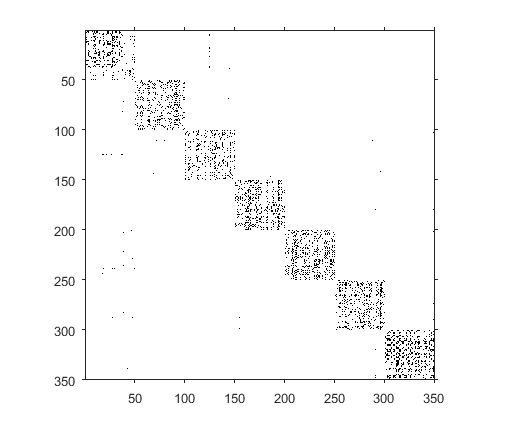}\label{fig:f32}}

	\caption{The sparse coefficient matrices for ROGSURE.}
\end{figure}

Given that our UoS structure for the modalities are jointly obtained in this case, the cross-sensor consistency is achieved by their intersection and hence by a product of the two sparse binary matrices $\mathbf{W}(1)$ and $\mathbf{W}(2)$ to produce one sparse matrix for both modalities, $\mathbf{W}_{total}$ resulting in an improved performance. Similar to ANDing process, the coefficient matrices from RoGSuRe are expected to share the same support and therefore their multiplication should yield a more reliable results. This is to be contrasted to the RoSuRe-based fusion when a weighted average for the sparse coefficient matrices worked much better than in that context since the data modalities might have different support so multiplication might lead to losing the unshared information.

\begin{equation}
\mathbf{W}_{Total}=\mathbf{W}(1).*\mathbf{W}(2)
\end{equation}

 The overall sparse matrix, $\mathbf{W}_{total}$ is displayed in Fig. 3(c). Observations belonging to one car are clustered as one subspace in which the contribution of each sensor is embedded in the entries of the $\mathbf{W}_{total}$. Using $\mathbf{W}_{total}$, in clustering proceeded in the same way as previously pointed out in II.B.1. As a result, the classification accuracy improved to 98.9\% as highlighted in Table 3.


\subsubsection{Experimental Validation with RoSuRe and RoGSuRe} \mbox{}\\
After learning the structure of the data clusters, we validate our results on the test data. We first extract the principal components (eigen vectors of the covariance matrix) of each cluster in the original (training) dataset. We subsequently project each new test point onto the subspace corresponding to each cluster, spanned by its principal components. The $l_2$ norm of the projection is then computed, and the class with the largest norm is selected to correspond to the class of this test point. We use the coefficient matrix $\mathbf{W}_{total}$ to cluster the test data points for both magnetometer and audio data. Classification on the new test data is jointly performed for both data modalities. The simulation results are listed in Table 3. From the results, it is clear that the Robust Group Subspace Recovery technique for the fused data remarkably outperforms RoSuRe.

\begin{table}[htbp]
	\caption{The validation performance for RoGSuRe and RoSuRe}
	\begin{center}
		
		\begin{tabular}{|c|c|c|c|c|}
			\hline
			
			&Learning &  Validation   \\
			
			\hline 
			
			RoGSuRe & 99.14\%  & 96.94\% \\ 
			\hline
			RoSuRe & 98.29\%  & 94.82\% \\ 
			\hline 
			
		\end{tabular}

	\end{center}
\end{table}

In the following, we will compare the performance of RoGSuRe to some existing and known multimodal subspace clustering technique such as LTMSC\cite{b64}, kMLRR \cite{b65}, kMSSC \cite{b65}, MSSC \cite{b65} and MLRR \cite{b65}. The reason our method outperforms the other methods is due to the group sparse term which does not enforce the similar structure across different modalities. Basically, the group-sparse term encourages different data modalities to communicate and share common information while at the same time each data modality maintains the relations between data points. The corresponding confusion matrices are dispalyed in Fig. 4. Multimodal RoSuRe algorithm separately considers each view and ignores the correlation that might exist among different views. If we consider the case of low quality data modalities, that might not share much commonality among their subspace structures, this will corrupt the support of the overall representation matrix, reduce the overall signal to noise ratio and  dramatically degrade the performance. Similarly, MLRR, MSSC, KMSSC and KMLRR, will be negatively impacted in case of low-quality data modalities since they enforce the same structure among different data views. On the other hand, RoGSuRe will be minimally affected in this case, as it provides a T-factor (assuming T modalities) improvement by allowing modalities to strengthen repeated relations. In particular, when addressing a  large number of modalities, the clustering improvement will be significant, and the gap between RoGSuRe and other approaches performance will be substantial. Other techniques, such as LTMSC, minimize the convex combination of the nuclear norms of all subspace representation matrices by seeking the lowest rank of the self-representation via a joint collaboration of multiple views. It, however, does not seem to provide richer information than unimodal data for our dataset. 

\begin{table}[htbp]
	\caption{Performance for different Multi-modal Subspace Clustering Methods}
	\begin{center}
		
		\begin{tabular}{|c|c|}
			\hline
			
			Method & Accuracy   \\
			
			\hline 
			
			LTMSC & 75.43\%   \\ 
			\hline
			KMLRR & 77.71\%   \\ 
			\hline 
			KMSSC & 79.14\%   \\ 
			
			\hline 
			MLRR & 89.14\%   \\ 
			\hline
			MSSC & 98.29\%   \\ 
			\hline
			RoSuRE & 98.29\%   \\ 
			\hline 
			\textbf{RoGSuRe} & \textbf{99.14}\%   \\ 
			\hline 
			
		\end{tabular}
		
		\label{tab4}
	\end{center}
\end{table}

\begin{figure}[htb]
	\centering
	\subfloat[Confusion matrix for LTMSC. ]{\includegraphics[width=4.4cm,height=3cm]{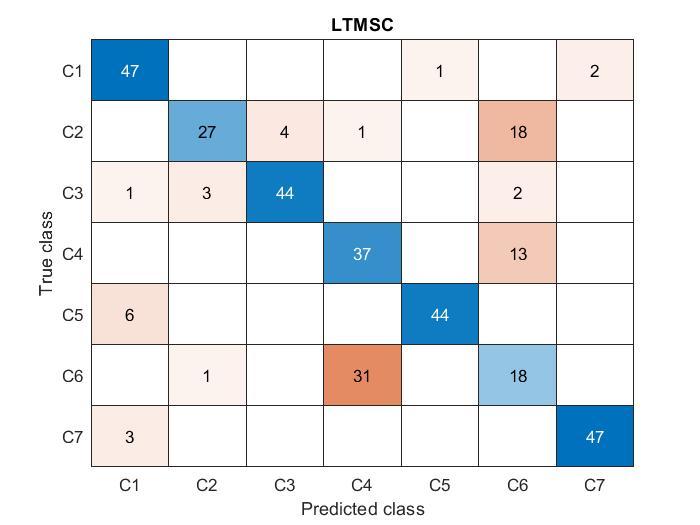}\label{fig:f40}}
	\hfill
	\subfloat[Confusion matrix for KMLRR. ]{\includegraphics[width=4.4cm,height=3cm]{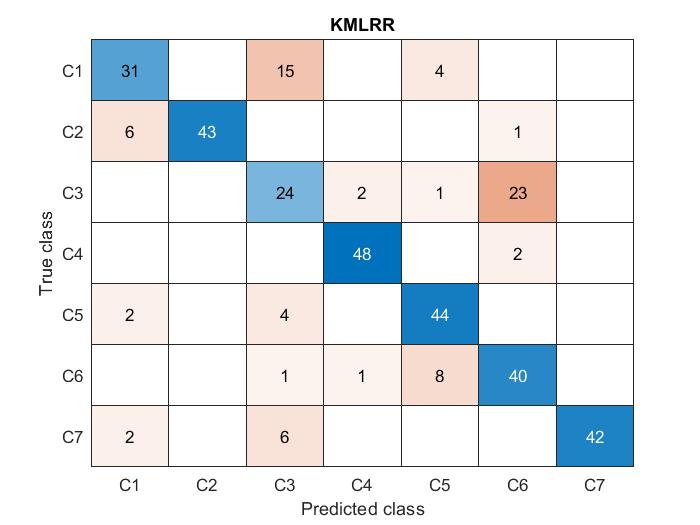}\label{fig:f41}}
	
	\centering
	\subfloat[Confusion matrix for KMSSC. ]{\includegraphics[width=4.4cm,height=3cm]{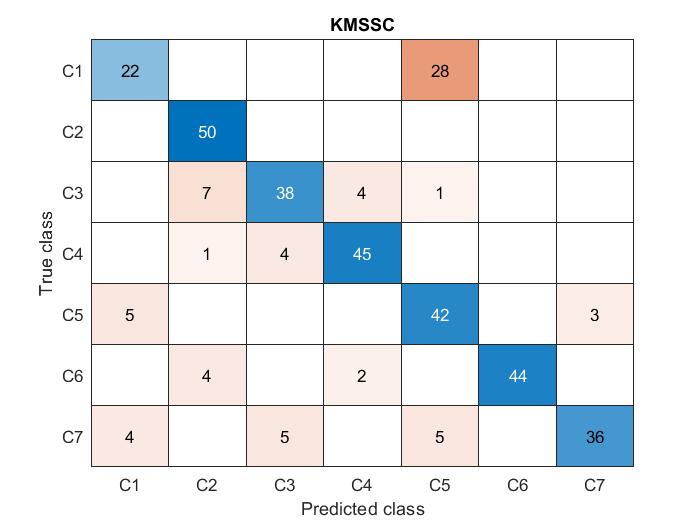}\label{fig:f42}}
	\hfill
	\subfloat[Confusion matrix for MLRR. ]{\includegraphics[width=4.4cm,height=3cm]{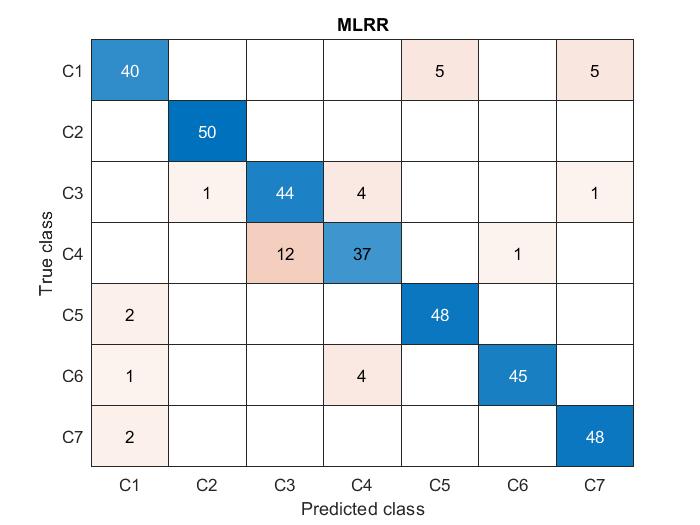}\label{fig:f43}}
	
	\centering
	\subfloat[Confusion matrix for MSSC. ]{\includegraphics[width=4.4cm,height=3cm]{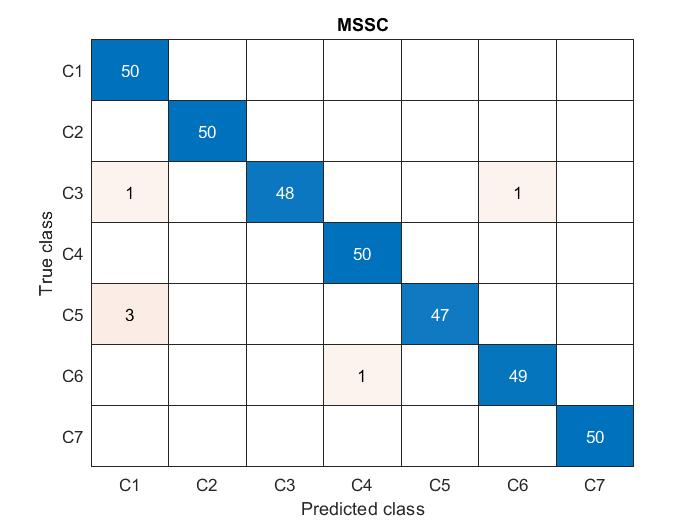}\label{fig:f44}}
	\hfill
	\subfloat[Confusion matrix for RoSuRe. ]{\includegraphics[width=4.4cm,height=3cm]{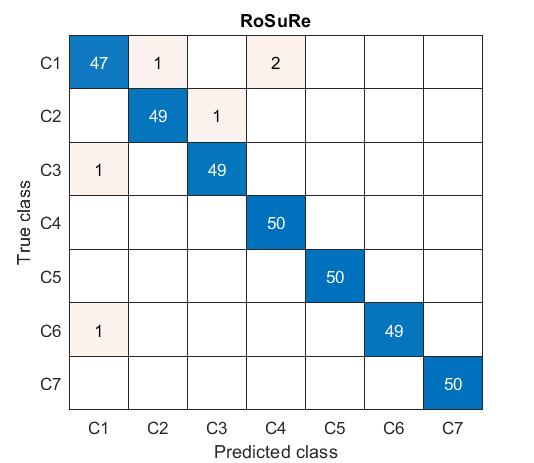}\label{fig:f45}}
	\centering
	\hfill
	\subfloat[Confusion matrix for RoGSuRe. ]{\includegraphics[width=4.4cm,height=3cm]{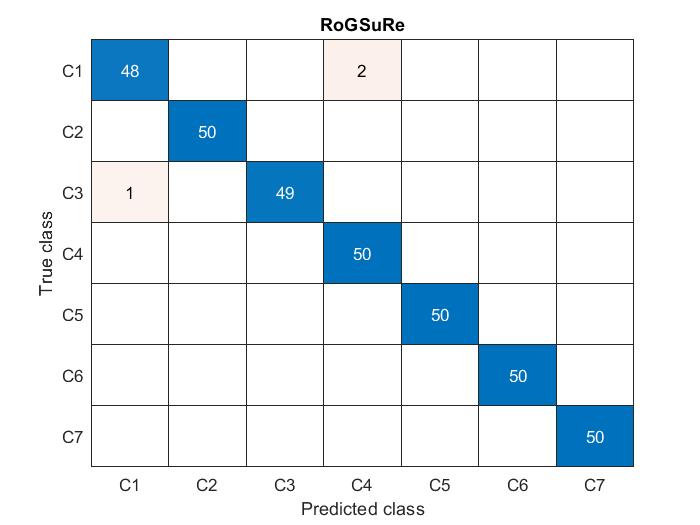}\label{fig:f46}}

	\caption{The confusion matrices for different multi-modal subspace clustering methods.}
\end{figure}

\section{Conclusion}

In this paper, we proposed two different approaches to fuse passive signal measured by low power instruments. The proposed approach recovers the underlying subspaces of data samples from measured data possibly corrupted by sparse errors. The multi-modal RoSuRe method is used to reliably and separately recover the subspace for each modality while the RoGSuRe manages to jointly optimize the subspace cluster structure. Both approaches provide a natural way to fuse multi-modal data by employing the self-representation matrix as an embedding in a shared domain. Experiments on real data are presented to demonstrate the effectiveness of this newly proposed method in solving the problem of subspace fusion with sparsely corrupted unlabeled data. Experimental results show a significant improvement for RoGSuRe over other state of the art subspace clustering techniques.

\appendices
\section{Proof of Theorem 1}
\begin{proof}
 We recall that we only study for the case $\rho=0$ in Eqn. (9). We denote by $\mathbf{z}_j(t)$ the dual vector associated with the constraints in Eqn. (9). In the following, we construct a solution $\mathbf{w}$ and a dual variable $\mathbf{z}_j(t)$ satisfying the optimality conditions of Eqn. (9) and the subspace detection property in Definition 1. For this purpose, we introduce the following optimization problem,
\begin{equation}
\min \limits_{\mathbf{\{\tilde{w}_j(t)\}_{t=1}^T}} \sum_{l} \sqrt{\sum_{t=1}^{T} \tilde{w}^2_{l,j}(t)} \ \mathbf{s.t.} \ \mathbf{x_j}(t)=\mathbf{X}^\alpha_{-j}(t) \mathbf{\tilde{w}_j}(t)
\end{equation}
 where $\alpha$ indicates the subspace of the $j^{th}$ datapoint, $\mathbf{X}^\alpha_{-j}(t)$ is every point in $\mathbf{X}(t)$ from the $\alpha$ subspace except $\mathbf{x_j}(t)$ and $ \tilde{w}_{l,j}(t)$ denotes the $l^{th}$ element in $ \tilde{w}_j(t)$. With an abuse of notation, we also take $\mathbf{\tilde{w}}_j(t)$ to be the optimal solution for Eqn. (17). Moreover, we denote its corresponding optimal dual vector with the smallest $l_2$ norm by $\mathbf{\tilde{z}_j}(t)$. Hence, $\mathbf{\tilde{z}_j}(t) \in \mathrm{col} \{\mathbf{X^\alpha_{-j}}(t)\}$, where $\mathrm{col}$ represent the column space, since otherwise, the projection of $\mathbf{\tilde{z}_j}(t)$ onto $\mathrm{col} \{\mathbf{X^\alpha_{-j}}(t)\}$ serves as another dual vector with strictly smaller $l_2$ norm.  Then, the optimality condition of Eqn. (17) yields $(\mathbf{X}^\alpha_{-j}(t))^T \mathbf{\tilde{z}}_j(t) \in \partial \sum_{l} \sqrt{\sum_{t=1}^{T} \tilde{w}^2_{l,j}(t)} $ where $\partial$ denote the sub-differential set such that,
\begin{equation}
\begin{aligned}
((\mathbf{X}^\alpha_{-j}(t))^T \mathbf{\tilde{z}}_j(t))_l=\frac{\tilde{w}_{lj}(t)}{\sqrt{\sum_t \tilde{w}_{lj}(t)}}\ \mathbf{if} \sum_t \tilde{w}_{lj}(t) \neq 0, \\
\sqrt{\sum_{t=1}^T (\mathbf{(X}^\alpha_{-j}(t))^T \mathbf{\tilde{z}}_j(t))^2_l}\leq 1 \ \mathbf{if} \sum_t \tilde{w}_{lj}(t) = 0. \ 
\end{aligned}
\end{equation}
Now, we construct $\mathbf{w_j}(t)$ by appropriately appending zero entries to $\mathbf{\tilde{w}_j}(t)$ whenever the indices $i,j$ of datapoints correspond to samples from different subspaces and when $i=j$, since each sample can be represented by other samples from the same subspace and can't represent itself. Furthermore, we take $\mathbf{z_j}(t)=\mathbf{\tilde{z_j}}(t)$. In the following, we prove that $\mathbf{w_j}(t)$ and $\mathbf{z_j}(t)$ satisfy the optimality condition of Eqn. (9), hence being an optimal solution with subspace detection property. The optimality condition for Eqn. (9) can be written as:

\begin{equation}
\begin{aligned}
(\mathbf{X}(t)\mathbf{z}_j(t))_k=\frac{w_{kj}(t)}{\sqrt{\sum_t w_{kj}(t)}}\ \mathbf{if} \ w_{kj}(t) \neq 0 \ \text{for some}\ t,\\
\sum_{t=1}^T ((\mathbf{X}(t))^T \mathbf{z}_j(t))^2_k\leq 1   \ \mathbf{if} \ \sum_t w^2_{kj}(t) = 0.  \  
\end{aligned}
\end{equation}
It is simple to check that the conditions in Eqn. (19) for $k \in S_j$ are satisfied by the definition in Eqn. (18). Also, note that for $k \notin \mathbf{S_j}$, we have $\mathbf{w_{k,j}}(t)=0$. Therefore, in order to prove that the subspace detection property holds, it remains to check that $ \forall k \notin S_j$, we have
\begin{equation}
\sum_{t=1}^T \langle \mathbf{x}_k(t),\mathbf{z}_j(t) \rangle ^2\textless 1. 
\end{equation}
  To proceed, we observe that
\begin{equation}
\sum_{t=1}^T (\mathbf{x}_k^T(t)\mathbf{z}_j(t))^2=\sum_{t=1}^T \parallel \mathbf{x}_k^T(t) \parallel^2 \parallel \mathbf{z}_j(t) \parallel^2 \cos^2(\theta_{k,j}(t)),
\end{equation} 
where $\theta_{k.j}(t)$ is the angle between $\mathbf{x}_k(t)$ and $\mathbf{z}_j(t)$. Since the data is normalized, $\parallel \mathbf{x}_k^T(t) \parallel^2=1$. Then, we have
\begin{equation}
\sum_{t=1}^T \parallel \mathbf{z}_j(t) \parallel^2 \cos^2(\theta(t)) \leq \max_t \cos^2(\theta(t)) \sum_{t=1}^T  \parallel \mathbf{z}_j(t)\parallel^2.
\end{equation}
The polar of a convex body $C$ is given by,
\begin{equation}
C^o=\left \{ \mathbf{x} \in \mathbb{R}^n \mid \mathbf{x}^T \mathbf{c} \leq 1 \forall \mathbf{c} \in \mathbf{C}\right \}
\end{equation}
Therefore, we observe that the polar $P^o_{-j}$ of $P_{-j}$ is given by:

\begin{equation}
P_{-j}^o=\left\{ \left \{ y(t) \right \}_{t=1}^T :\sum_{t=1}^T(x_q(t)^Ty(t))^2 \leq 1\right\}
\end{equation}
As a result, from Eqn. (18), we conclude that $\left \{ \mathbf{z}_j(t) \right \}_{t=1}^T \in P_{-j}^o$. We define the circumradius of a convex subset $P$ of the finite dimension Euclidean space as the radius of the smallest Euclidean ball containing $P$ and denote it by $R(P)$. Hence,

\begin{equation}
\sum_{t=1}^T \parallel \mathbf{z_j}(t)\parallel^2 \leq  R^2(P_{-j}^o)  
\end {equation}

For a symmetric convex body $P$, the following relationship between the inradius of $P$ and circumradius of its polar $P^o$ holds \cite{b33}:
\begin{equation}
r(P)R(P^o)=1
\end{equation}

Therefore,
\begin{equation}
\sum_{t=1}^T \parallel \mathbf{z_j}(t)\parallel^2 \leq  \frac{1}{r^2(P_{-j})}
\end{equation}
In summary, Eqns. (22) and (26) imply that it suffices to verify that $\forall k \notin S_j$, we have,
\begin{equation}
\max_t \cos^2\theta (t) \textless \min_t  r^2(P_{-j})
\end{equation}
Then, the condition in Eqn. (20) is satisfied and $\mathbf{w}_j(t)$ is a solution for Eqn. (9) when $\rho=0$ which implies that the subspace detection property holds.

\end{proof}

\section*{Acknowledgment}

The work of the first three authors were in part supported by DOE-National Nuclear Security Administration through CNEC-NCSU under Award DE-NA0002576.

\ifCLASSOPTIONcaptionsoff
  \newpage
\fi

\bibliographystyle{IEEEtran}
\bibliography{paper_writeup_journal}

\begin{thebibliography}{10}
\providecommand{\url}[1]{#1}
\csname url@samestyle\endcsname
\providecommand{\newblock}{\relax}
\providecommand{\bibinfo}[2]{#2}
\providecommand{\BIBentrySTDinterwordspacing}{\spaceskip=0pt\relax}
\providecommand{\BIBentryALTinterwordstretchfactor}{4}
\providecommand{\BIBentryALTinterwordspacing}{\spaceskip=\fontdimen2\font plus
\BIBentryALTinterwordstretchfactor\fontdimen3\font minus
  \fontdimen4\font\relax}
\providecommand{\BIBforeignlanguage}[2]{{%
\expandafter\ifx\csname l@#1\endcsname\relax
\typeout{** WARNING: IEEEtran.bst: No hyphenation pattern has been}%
\typeout{** loaded for the language `#1'. Using the pattern for}%
\typeout{** the default language instead.}%
\else
\language=\csname l@#1\endcsname
\fi
#2}}
\providecommand{\BIBdecl}{\relax}
\BIBdecl

\bibitem{b38}
E.~J. Cand{\`e}s, X.~Li, Y.~Ma, and J.~Wright, ``Robust principal component
  analysis?'' \emph{Journal of the ACM (JACM)}, vol.~58, no.~3, p.~11, 2011.

\bibitem{b12}
M.~Elad, M.~A. Figueiredo, and Y.~Ma, ``On the role of sparse and redundant
  representations in image processing,'' \emph{Proceedings of the IEEE},
  vol.~98, no.~6, pp. 972--982, 2010.

\bibitem{b40}
V.~M. Patel and R.~Vidal, ``Kernel sparse subspace clustering,'' in \emph{2014
  IEEE International Conference on Image Processing (ICIP)}.\hskip 1em plus
  0.5em minus 0.4em\relax IEEE, 2014, pp. 2849--2853.

\bibitem{b41}
X.~Peng, S.~Xiao, J.~Feng, W.-Y. Yau, and Z.~Yi, ``Deep subspace clustering
  with sparsity prior.'' in \emph{IJCAI}, 2016, pp. 1925--1931.

\bibitem{b46}
A.~Y. Yang, S.~R. Rao, and Y.~Ma, ``Robust statistical estimation and
  segmentation of multiple subspaces,'' in \emph{2006 Conference on Computer
  Vision and Pattern Recognition Workshop (CVPRW'06)}.\hskip 1em plus 0.5em
  minus 0.4em\relax IEEE, 2006, pp. 99--99.

\bibitem{b44}
W.~Hong, J.~Wright, K.~Huang, and Y.~Ma, ``Multiscale hybrid linear models for
  lossy image representation,'' \emph{IEEE Transactions on Image Processing},
  vol.~15, no.~12, pp. 3655--3671, 2006.

\bibitem{b45}
R.~Vidal, S.~Soatto, Y.~Ma, and S.~Sastry, ``An algebraic geometric approach to
  the identification of a class of linear hybrid systems,'' in \emph{42nd IEEE
  International Conference on Decision and Control (IEEE Cat. No. 03CH37475)},
  vol.~1.\hskip 1em plus 0.5em minus 0.4em\relax IEEE, 2003, pp. 167--172.

\bibitem{b56}
W.~Tang, A.~Panahi, H.~Krim, and L.~Dai, ``Analysis dictionary learning based
  classification: Structure for robustness,'' \emph{IEEE Transactions on Image
  Processing}, vol.~28, no.~12, pp. 6035--6046, 2019.

\bibitem{b16}
X.~Bian, A.~Panahi, and H.~Krim, ``Bi-sparsity pursuit: A paradigm for robust
  subspace recovery,'' \emph{Signal Processing}, 2018.

\bibitem{b47}
D.~Lahat, T.~Adali, and C.~Jutten, ``Multimodal data fusion: an overview of
  methods, challenges, and prospects,'' \emph{Proceedings of the IEEE}, vol.
  103, no.~9, pp. 1449--1477, 2015.

\bibitem{b50}
R.~K. Sharma, ``Probabilistic model-based multisensor image fusion,'' Ph.D.
  dissertation, Ph. D.), Oregon Graduate Institute, 1999.

\bibitem{b51}
O.~Hellwich and C.~Wiedemann, ``Object extraction from high-resolution
  multisensor image data,'' in \emph{Third International Conference Fusion of
  Earth Data, Sophia Antipolis}, vol. 115, 2000.

\bibitem{b52}
Z.~Korona and M.~M. Kokar, ``Model theory based fusion framework with
  application to multisensor target recognition,'' in \emph{1996 IEEE/SICE/RSJ
  International Conference on Multisensor Fusion and Integration for
  Intelligent Systems (Cat. No. 96TH8242)}.\hskip 1em plus 0.5em minus
  0.4em\relax IEEE, 1996, pp. 9--16.

\bibitem{b54}
F.~K. Soong and A.~E. Rosenberg, ``On the use of instantaneous and transitional
  spectral information in speaker recognition,'' \emph{IEEE Transactions on
  Acoustics, Speech, and Signal Processing}, vol.~36, no.~6, pp. 871--879,
  1988.

\bibitem{b53}
L.~Xu, A.~Krzyzak, and C.~Y. Suen, ``Methods of combining multiple classifiers
  and their applications to handwriting recognition,'' \emph{IEEE transactions
  on systems, man, and cybernetics}, vol.~22, no.~3, pp. 418--435, 1992.

\bibitem{b2}
A.~Khosravi, E.~Mazloumi, S.~Nahavandi, D.~Creighton, and J.~Van~Lint,
  ``Prediction intervals to account for uncertainties in travel time
  prediction,'' \emph{IEEE Transactions on Intelligent Transportation Systems},
  vol.~12, no.~2, pp. 537--547, 2011.

\bibitem{b4}
Y.~A. Kathawala and B.~Tueck, ``The use of rfid for traffic management,''
  \emph{International journal of technology, policy and management}, vol.~8,
  no.~2, pp. 111--125, 2008.

\bibitem{b6}
R.~O. Sanchez, C.~Flores, R.~Horowitz, R.~Rajagopal, and P.~Varaiya, ``Vehicle
  re-identification using wireless magnetic sensors: Algorithm revision,
  modifications and performance analysis,'' in \emph{Vehicular Electronics and
  Safety (ICVES), 2011 IEEE International Conference on}.\hskip 1em plus 0.5em
  minus 0.4em\relax IEEE, 2011, pp. 226--231.

\bibitem{b7}
A.~Haoui, R.~Kavaler, and P.~Varaiya, ``Wireless magnetic sensors for traffic
  surveillance,'' \emph{Transportation Research Part C: Emerging Technologies},
  vol.~16, no.~3, pp. 294--306, 2008.

\bibitem{b8}
C.~T. Christou and G.~M. Jacyna, ``Vehicle detection and localization using
  unattended ground magnetometer sensors,'' in \emph{Information Fusion
  (FUSION), 2010 13th Conference on}.\hskip 1em plus 0.5em minus 0.4em\relax
  IEEE, 2010, pp. 1--8.

\bibitem{b9}
H.~Krim and M.~Viberg, ``Two decades of array signal processing research: the
  parametric approach,'' \emph{IEEE signal processing magazine}, vol.~13,
  no.~4, pp. 67--94, 1996.

\bibitem{b10}
Y.~Ding, B.~Banitalebi, T.~Miyaki, and M.~Beigl, ``Rftraffic: a study of
  passive traffic awareness using emitted rf noise from the vehicles,''
  \emph{EURASIP Journal on Wireless Communications and Networking}, vol. 2012,
  no.~1, p.~8, 2012.

\bibitem{b59}
L.~Li, Z.-Q. Luo, K.~M. Wong, and E.~Bosse, ``Convex optimization approach to
  identify fusion for multisensor target tracking,'' \emph{IEEE Transactions on
  Systems, Man, and Cybernetics-Part A: Systems and Humans}, vol.~31, no.~3,
  pp. 172--178, 2001.

\bibitem{b60}
M.~C. Florea and E.~Bosse, ``Critiques on some combination rules for
  probability theory based on optimization techniques,'' in \emph{2007 10th
  International Conference on Information Fusion}.\hskip 1em plus 0.5em minus
  0.4em\relax IEEE, 2007, pp. 1--8.

\bibitem{b61}
L.~Li, ``Data fusion and filtering for target tracking and identification,''
  Ph.D. dissertation, 2003.

\bibitem{b57}
D.~L. Hall and J.~Llinas, ``An introduction to multisensor data fusion,''
  \emph{Proceedings of the IEEE}, vol.~85, no.~1, pp. 6--23, 1997.

\bibitem{b58}
B.~Khaleghi, A.~Khamis, F.~O. Karray, and S.~N. Razavi, ``Multisensor data
  fusion: A review of the state-of-the-art,'' \emph{Information fusion},
  vol.~14, no.~1, pp. 28--44, 2013.

\bibitem{b37}
S.~Ghanem, A.~Panahi, H.~Krim, R.~A. Kerekes, and J.~Mattingly, ``Information
  subspace-based fusion for vehicle classification,'' in \emph{2018 26th
  European Signal Processing Conference (EUSIPCO)}.\hskip 1em plus 0.5em minus
  0.4em\relax IEEE, 2018, pp. 1612--1616.

\bibitem{b62}
B.~Cheng, G.~Liu, J.~Wang, Z.~Huang, and S.~Yan, ``Multi-task low-rank affinity
  pursuit for image segmentation,'' in \emph{2011 International Conference on
  Computer Vision}.\hskip 1em plus 0.5em minus 0.4em\relax IEEE, 2011, pp.
  2439--2446.

\bibitem{b39}
G.~Liu, Z.~Lin, S.~Yan, J.~Sun, Y.~Yu, and Y.~Ma, ``Robust recovery of subspace
  structures by low-rank representation,'' \emph{IEEE transactions on pattern
  analysis and machine intelligence}, vol.~35, no.~1, pp. 171--184, 2012.

\bibitem{b64}
C.~Zhang, H.~Fu, S.~Liu, G.~Liu, and X.~Cao, ``Low-rank tensor constrained
  multiview subspace clustering,'' in \emph{Proceedings of the IEEE
  international conference on computer vision}, 2015, pp. 1582--1590.

\bibitem{b65}
M.~Abavisani and V.~M. Patel, ``Multimodal sparse and low-rank subspace
  clustering,'' \emph{Information Fusion}, vol.~39, pp. 168--177, 2018.

\bibitem{b63}
E.~Elhamifar and R.~Vidal, ``Sparse subspace clustering: Algorithm, theory, and
  applications,'' \emph{IEEE transactions on pattern analysis and machine
  intelligence}, vol.~35, no.~11, pp. 2765--2781, 2013.

\bibitem{b17}
Z.~Lin, R.~Liu, and Z.~Su, ``Linearized alternating direction method with
  adaptive penalty for low-rank representation,'' in \emph{Advances in neural
  information processing systems}, 2011, pp. 612--620.

\bibitem{b25}
A.~Y. Ng, M.~I. Jordan, and Y.~Weiss, ``On spectral clustering: Analysis and an
  algorithm,'' in \emph{Advances in neural information processing systems},
  2002, pp. 849--856.

\bibitem{b31}
D.~Taylor, S.~Shai, N.~Stanley, and P.~J. Mucha, ``Enhanced detectability of
  community structure in multilayer networks through layer aggregation,''
  \emph{Physical review letters}, vol. 116, no.~22, p. 228301, 2016.

\bibitem{b32}
M.~Soltanolkotabi, E.~J. Candes \emph{et~al.}, ``A geometric analysis of
  subspace clustering with outliers,'' \emph{The Annals of Statistics},
  vol.~40, no.~4, pp. 2195--2238, 2012.

\bibitem{b34}
E.~Elhamifar and R.~Vidal, ``Sparse subspace clustering,'' in \emph{2009 IEEE
  Conference on Computer Vision and Pattern Recognition}.\hskip 1em plus 0.5em
  minus 0.4em\relax IEEE, 2009, pp. 2790--2797.

\bibitem{b35}
------, ``Clustering disjoint subspaces via sparse representation,'' in
  \emph{2010 IEEE International Conference on Acoustics, Speech and Signal
  Processing}.\hskip 1em plus 0.5em minus 0.4em\relax IEEE, 2010, pp.
  1926--1929.

\bibitem{b36}
R.~A. Kerekes, T.~P. Karnowski, M.~Kuhn, M.~R. Moore, B.~Stinson, R.~Tokola,
  A.~Anderson, and J.~M. Vann, ``Vehicle classification and identification
  using multi-modal sensing and signal learning,'' in \emph{2017 IEEE 85th
  Vehicular Technology Conference (VTC Spring)}.\hskip 1em plus 0.5em minus
  0.4em\relax IEEE, 2017, pp. 1--5.

\bibitem{b18}
B.~Malhotra, I.~Nikolaidis, and J.~Harms, ``Distributed classification of
  acoustic targets in wireless audio-sensor networks,'' \emph{Computer
  Networks}, vol.~52, no.~13, pp. 2582--2593, 2008.

\bibitem{b20}
J.~Kell, I.~Fullerton, and M.~Mills, ``Traffic detector handbook. federal
  highway administration, vol. i,'' FHWA-HRT-06-108, October, Tech. Rep., 2006.

\bibitem{b21}
S.~B. Davis and P.~Mermelstein, ``Comparison of parametric representations for
  monosyllabic word recognition in continuously spoken sentences,'' in
  \emph{Readings in speech recognition}.\hskip 1em plus 0.5em minus 0.4em\relax
  Elsevier, 1990, pp. 65--74.

\bibitem{b23}
S.~Charbonnier, A.-C. Pitton, and A.~Vassilev, ``Vehicle re-identification with
  a single magnetic sensor,'' in \emph{Instrumentation and Measurement
  Technology Conference (I2MTC), 2012 IEEE International}.\hskip 1em plus 0.5em
  minus 0.4em\relax IEEE, 2012, pp. 380--385.

\bibitem{b24}
I.~T. Jolliffe, ``A note on the use of principal components in regression,''
  \emph{Applied Statistics}, pp. 300--303, 1982.

\bibitem{b26}
E.~W. Forgy, ``Cluster analysis of multivariate data: efficiency versus
  interpretability of classifications,'' \emph{biometrics}, vol.~21, pp.
  768--769, 1965.

\bibitem{b28}
Z.~Zivkovic, ``Improved adaptive gaussian mixture model for background
  subtraction,'' in \emph{Pattern Recognition, 2004. ICPR 2004. Proceedings of
  the 17th International Conference on}, vol.~2.\hskip 1em plus 0.5em minus
  0.4em\relax IEEE, 2004, pp. 28--31.

\bibitem{b29}
D.~A. Reynolds, T.~F. Quatieri, and R.~B. Dunn, ``Speaker verification using
  adapted gaussian mixture models,'' \emph{Digital signal processing}, vol.~10,
  no. 1-3, pp. 19--41, 2000.

\bibitem{b30}
L.~Rokach and O.~Maimon, ``Clustering methods,'' in \emph{Data mining and
  knowledge discovery handbook}.\hskip 1em plus 0.5em minus 0.4em\relax
  Springer, 2005, pp. 321--352.

\bibitem{b33}
R.~Brandenberg, A.~Dattasharma, P.~Gritzmann, and D.~Larman, ``Isoradial
  bodies,'' \emph{Discrete \& Computational Geometry}, vol.~32, no.~4, pp.
  447--457, 2004.

\end{thebibliography}

\begin{IEEEbiographynophoto}{Sally Ghanem}
	received the B.Sc. degree in electrical engineering from Alexandria University, in 2013, and the M.Sc. degree in electrical and computer engineering from North Carolina State University, Raleigh, NC, USA in 2016, where she is currently pursuing the Ph.D. degree. During the PhD program, she has spent time at Oak Ridge National Laboratory working on multimodal data. Her research interests include the areas of computer vision, digital signal processing, image processing and machine learning. 
\end{IEEEbiographynophoto}

\begin{IEEEbiographynophoto}{Ashkan Panahi}
	received the B.Sc. degree from the Iran University of Science and Technology, Tehran, Iran, in 2007, the M.Sc. degree in communication systems engineering and electrical engineering from Chalmers University, in 2010, and the Ph.D. degree in signal processing from the Electrical Engineering Department, Chalmers University, Sweden, in 2015. Furthermore, he has held a post-doctoral research position at the Computer Science Department, Chalmers University, where he is currently an Assistant Professor. He has been a visiting student at the Electrical Engineering Department, California Institute of Technology (Caltech). He was also a U.S. National Research Council Post-Doctoral Researcher with North Carolina State University. His research includes the development and application of optimization algorithms for various signal processing and machine learning tasks for a large amount of data, especially computer vision and image processing.
\end{IEEEbiographynophoto}

\begin{IEEEbiographynophoto}{Hamid Krim}
	received the B.Sc. and M.Sc. and Ph.D. in ECE. He was a Member of Technical Staff at AT\&T Bell Labs, where he has conducted research and development in the areas of telephony and digital communication systems/subsystems. Following an NSF Postdoctoral Fellowship at Foreign Centers of Excellence, LSS/University of Orsay, Paris, France, he joined the Laboratory for Information and Decision Systems, MIT, Cambridge, MA, USA, as a Research Scientist and where he performed/supervised research. He is currently a Professor of electrical engineering in the Department of Electrical and Computer Engineering, North Carolina State University, NC, leading the Vision, Information, and Statistical Signal Theories and Applications Group. His research interests include statistical signal and image analysis and mathematical modeling with a keen emphasis on applied problems in classification and recognition using geometric and topological tools. He has served on the SP society Editorial Board and on TCs, and is the SP Distinguished Lecturer for 2015-2016.
\end{IEEEbiographynophoto}

\begin{IEEEbiographynophoto}{Ryan A. Kerekes}
	received the B.S. degree in computer engineering from the University of Tennessee, Knoxville, TN, USA, in 2003, and the M.S. and Ph.D. degrees in electrical and computer engineering from Carnegie Mellon University, Pittsburgh, PA, USA, in 2005 and 2007, respectively.
	He currently leads the RF Communications and Intelligent Systems group, Oak Ridge National Laboratory, Oak Ridge, TN, USA. His research interests include the areas of signal and image processing and machine learning for national security applications.
\end{IEEEbiographynophoto}

\end{document}